\documentclass[journal]{IEEEtran}
\usepackage{amsmath,amssymb}
\usepackage{booktabs}
\usepackage{cite}
\usepackage{graphicx}
\usepackage{multirow}
\usepackage{xcolor}
\usepackage{url}
\usepackage{hyperref}

\usepackage{amsmath,amsfonts,bm}

\def\eqref#1{equation~\ref{#1}}

\def\1{\bm{1}}

\DeclareMathAlphabet{\mathsfit}{\encodingdefault}{\sfdefault}{m}{sl}
\SetMathAlphabet{\mathsfit}{bold}{\encodingdefault}{\sfdefault}{bx}{n}

\graphicspath{{./}{figure/}{../figure/}}

\newcommand{\Rtwo}{R^2}

\title{Textual Environmental Context and Spatial Graphs for LLM-Based Regional SST Forecasting}

\author{Xiong~Li, Xiaowei~Zhou, Yanwei~Yu, Qian~Cui, and Junyu~Dong%
\thanks{All authors are with Ocean University of China, Qingdao, China.}}

\begin{document}

\maketitle

\begin{abstract}
Sea surface temperature (SST) forecasting depends on local temporal persistence,
regional spatial dependence, and environmental conditions that evolve with the
forecast date. We study how these heterogeneous conditions can be presented to a
large language model (LLM) for regional multi-step forecasting without serializing
the full SST grid as text. We formulate forecasting as conditional numerical
generation: historical SST and anomaly sequences, date-aligned environmental
records, and static ocean knowledge form a textual context, while regional spatial
state is supplied through continuous graph-derived prefixes. A static graph encodes
persistent geographic--climatological relations, and a dynamic graph encodes recent
SST correlations and localized tropical-cyclone influence. Two graph neural
networks produce a target-node representation that is mapped by a spatial-prefix fusion and injected into the LLM input. On SST forecasting in
the South China Sea, the complete configuration achieves the best MAE and
$\Rtwo$ among the compared methods over ten forecast steps. Alongside the
numerical forecast, a rule-based module matches predicted trends and
environmental-factor directions with knowledge entries to return source-linked,
post-hoc contextual explanations.
\end{abstract}

\section{Introduction}
\label{sec:introduction}

Sea surface temperature (SST) forecasting supports marine environmental
monitoring and related applications~\cite{liu2024sstseasonal,cui2025wenhai}.
SST evolution reflects local temporal
persistence, spatial dependence among ocean cells, and environmental conditions
such as tropical cyclones, large-scale climate states, and regional
circulation~\cite{england2025heatwave}.
A forecasting model that uses only the historical SST sequence of one target
cell must infer these heterogeneous influences indirectly. This motivates a
central question: how can date-aligned environmental context and regional
spatial state be presented jointly to a large language model (LLM) for
multi-step SST forecasting?

Recent work has explored the use of LLMs for numerical and text-conditioned forecasting~\cite{zhong2025timevlm,wu2026aurora}. From News
to Forecast combines selected news, auxiliary information, and historical
sequences in textual prompts for numerical generation~\cite{wang2024news}. In
the SST domain, OKG-LLM aligns
ocean-domain knowledge representations with SST observations through an
LLM-based prediction pipeline~\cite{yang2026okgllm}. These studies establish
that LLMs can incorporate contextual information and domain knowledge into
numerical forecasting. However, they do not directly address how
forecast-date environmental descriptions and explicit, date-varying regional
spatial relations should be jointly supplied to an LLM for regional multi-step SST forecasting.

We address this problem by assigning textual and spatial conditions distinct
roles. Related studies have paired auxiliary context with numerical histories,
aligned time-series representations with language, and adapted models to
heterogeneous covariates~\cite{liu2024timemmd,hu2025contextalignment,han2026unica}.
Tropical-cyclone records are organized as descriptions of time, location,
and intensity; the ENSO index and regional circulation provide climate and
atmospheric context; and static ocean knowledge describes mixing, advection,
upwelling, and air--sea exchange. These sources form a textual forecasting
context together with the historical SST and anomaly sequences of the target
cell. Text, however, does not itself perform neighborhood feature aggregation
over the SST field: cells with similar environmental descriptions may have
different local neighborhood states or storm influences. We therefore represent
persistent and date-varying spatial relations separately. A static graph encodes
geographic proximity and climatological features, whereas a dynamic graph
encodes recent SST correlations and localized tropical-cyclone influence.

The two GNN branches aggregate regional information and produce a
representation for the target cell. Rather than serializing all grid cells as
text, a factorized spatial-prefix projector maps this representation into a
compact sequence of continuous prefix tokens, which is concatenated with the
LLM input embeddings for autoregressive SST generation. Alongside the numerical forecast, a separate rule-based module matches predicted trends and
environmental-factor directions with knowledge entries to return source-linked post-hoc contextual explanations.

The main contributions are as follows:
\begin{itemize}
    \item \textbf{A time-aligned text--graph interface for regional SST forecasting.}
    We formulate multi-step SST forecasting as conditional numerical generation,
    where environmental records and ocean knowledge are represented as text and
    regional spatial state is supplied through continuous graph-derived prefixes.

    \item \textbf{Date-specific dual-graph spatial conditioning.}
    We combine a static geographic--climatological graph with a dynamic graph
    constructed from recent SST correlations and localized tropical-cyclone
    influence, then inject the target-node state into the LLM without increasing
    the input length with the number of grid cells.

    \item \textbf{Lead-wise empirical evaluation of text and spatial conditions.}
    On South China Sea SST forecasting, we report lead-wise comparisons and
    ablations. Among the compared baseline methods, the full configuration achieves the best performance on average.
\end{itemize}

\section{Related Work}
\label{sec:related_work}

\subsection{Spatiotemporal SST forecasting}

Recent SST forecasting studies have improved spatiotemporal modeling through
spatial-relation learning, motion features, and physical priors. ASTGRN learns
adaptive node representations and graph structures for sea-area dependencies,
whereas DDSTGNN separates diffusion and intrinsic signals and learns evolving
spatiotemporal dependencies with a dynamic graph~\cite{zhao2024astgrn,jin2025ddstgnn}.
Peng \textit{et al.} combine optical-flow guidance with SwinLSTM features to
extract motion information from SST fields~\cite{peng2025dualpath}. These
methods provide different structural inductive biases for regional variation.

Physics-guided methods combine data-driven representations with ocean-process
priors. DP-BICNN couples physical and data-driven processes through bidirectional
information compensation, while PCFNet combines frequency-domain features with
an advection--diffusion constraint~\cite{liu2025dpbicnn,liu2026pcfnet}.
PG-DSBN combines numerical simulation priors with deterministic--stochastic
modeling, and physics-constrained adapter tuning transfers meteorological
foundation representations to global SST forecasting~\cite{niu2026pgdsbn,li2026physicsadapter}.
These studies provide strong numerical spatiotemporal and physics-guided
inductive biases for SST prediction. The complementary question studied here is how forecast-date environmental descriptions and a compact representation of regional spatial state can be jointly supplied to an LLM for conditional
numerical generation. Our method represents the latter through static and
dynamic graphs rather than imposing an explicit physical-equation constraint.

\subsection{LLMs for numerical and text-conditioned forecasting}

Some work adapt time-series to the representation space of language
models. Time-LLM aligns time-series patches with text prototypes and uses a
prompt prefix to provide task context~\cite{jin2024timellm}. TimeCMA aligns
numerical and prompt encodings, LangTime combines temporal-comprehension prompts
with reinforcement-learning fine-tuning, and SE-LLM introduces temporal--semantic
interaction and a Time-Adapter~\cite{liu2025timecma,niu2025langtime,liu2026sellm}.
These methods primarily address the interface between numerical patterns and
language-model representations.

Another line studies the predictive value of textual context. From News to
Forecast integrates selected news with historical sequences and fine-tunes an
LLM for numerical generation~\cite{wang2024news}. VoT jointly uses exogenous
event text and endogenous sequence descriptions, with fusion at both the
representation and prediction-frequency levels; TaTS treats time-aligned text
as an auxiliary variable for numerical forecasting~\cite{wang2026vot,li2026tats}.
For the related task of text-conditioned sequence synthesis, VerbalTS generates
time series from unstructured descriptions~\cite{gu2025verbalts}. Our setting
focuses on regional SST forecasting, where date-aligned environmental records
and static ocean knowledge form textual conditions, while a separate spatial
branch summarizes regional SST state for the target cell.

\subsection{Graph-conditioned inputs for language models}

Graph convolution learns node representations by aggregating neighborhood
information, providing a basis for regional spatial modeling~\cite{kipf2017gcn}.
When graph representations are supplied to an LLM, the central interface problem
is converting non-sequential structure into token-level inputs. GraphToken learns
graph-conditioned soft prompts from GNN representations, while G-Retriever
combines subgraph retrieval, graph encoding, and soft prompting for textual-graph
question answering~\cite{perozzi2024graphtoken,he2024gretriever}. GOFA inserts
GNN layers into an LLM for joint graph--language modeling, and LGPT studies
query-aware learnable graph-pooling tokens to balance token cost and information
compression~\cite{kong2025gofa,kim2025lgpt}.

These studies motivate the use of continuous graph-derived tokens as an
interface between graph encoders and LLMs. Our setting differs from graph
question answering and graph-language modeling: static and dynamic GNNs first
aggregate regional SST information, and the resulting target-node state is
projected into continuous prefix tokens for conditional numerical generation.
The representation is conditioned on the forecast date and target location,
while its prefix length remains independent of the number of regional grid
cells.

\subsection{Ocean knowledge and forecast interpretation}

OKG-LLM aligns ocean-knowledge representations with SST observations through an
LLM-based prediction pipeline~\cite{yang2026okgllm}. In contrast, our framework
uses static ocean knowledge and forecast-date environmental records as textual
conditions, while encoding regional spatial state with target-specific prefixes
derived from static and dynamic graphs.

TimeCAP, TimeXL, and TS-RAG incorporate context generation, refinement, or
retrieval into time-series prediction pipelines~\cite{lee2025timecap,
jiang2025timexl,ning2025tsrag}. Our knowledge lookup runs separately after
forecasting, matching predicted trends and environmental-factor directions to
source-linked entries for rule-based post-hoc contextualization.

\section{Method}
\label{sec:method}

\subsection{Task formulation and framework overview}
\label{sec:task}

Let the study region contain $N$ regular grid cells. Cell $i$ has coordinates
$\mathbf c_i$ and an ocean mask $m_i\in\{0,1\}$. At forecast origin $t$, let
$\mathbf X_t=[\mathbf x_{t-L+1},\ldots,\mathbf x_t]\in\mathbb R^{L\times N}$
and $\mathbf D_t=[\mathbf d_{t-L+1},\ldots,\mathbf d_t]\in\mathbb R^{L\times N}$
denote the regional SST and anomaly histories, respectively. The
forecast-date environmental record $\mathbf e_t$ contains tropical-cyclone,
ENSO, and regional-circulation context, while $K$ denotes static ocean
knowledge. All date-dependent environmental records and dynamic-graph statistics are
constructed using observations available no later than forecast origin $t$.

\begin{figure*}[t]
 \centering
 \includegraphics[width=\textwidth]{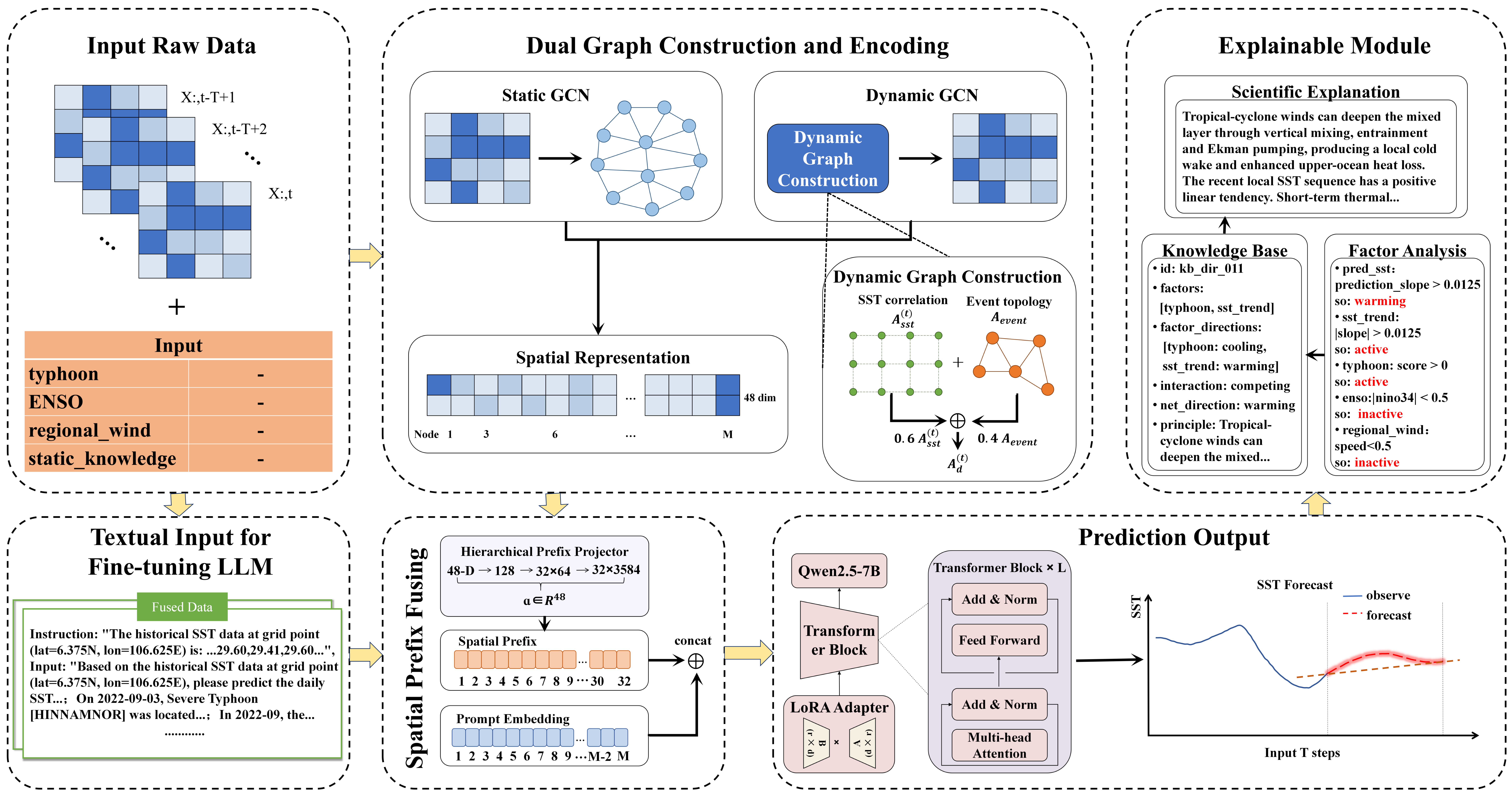}
 \caption{Overview of the framework. At forecast origin $t$, target-cell SST and anomaly
histories, environmental records, and static ocean knowledge form a textual
condition. Regional observations construct a static geographic--climatological
graph and a date-specific dynamic graph. Two GNN branches encode the target-node
state, which a factorized projector maps to continuous prefix tokens for
autoregressive LLM forecasting. A separate rule-based module then matches
forecast trends and environmental factors with knowledge entries for
source-linked post-hoc contextualization.}
 \label{fig:pipeline}
\end{figure*}

We construct a textual condition $q_{t,i}$ from the target-cell histories,
environmental record, and static knowledge, and construct a static graph
$\mathcal G^s=(\mathbf A^s,\mathbf F^s)$ and a date-specific dynamic graph
$\mathcal G^d_t=(\mathbf A^d_t,\mathbf F^d_t)$ for the regional SST field.
For an ocean target cell $i$, the task is to predict the next $H$ SST values,
$\mathbf y_{t,i}=(x_{t+1,i},\ldots,x_{t+H,i})$, as
\begin{equation}
 \widehat{\mathbf y}_{t,i}
 =f_\theta\!\left(q_{t,i},\mathcal G^s,\mathcal G^d_t,i\right).
 \label{eq:task}
\end{equation}
The model compresses the graph-derived target-node state into continuous prefix
tokens and concatenates them with the embeddings of $q_{t,i}$ for
autoregressive numerical generation. Each
sample corresponds to one forecast origin and one target cell; samples from the
same forecast origin share the regional graph inputs but receive target-specific
textual conditions and spatial prefixes. Land cells remain in the regular
indexing scheme for graph construction, but are masked and excluded from
prediction targets and evaluation. Figure.~\ref{fig:pipeline} summarizes the framework.

\subsection{Multi-source observations and textual input}
\label{sec:text_input}

\paragraph{Historical observations and environmental context.}
We use the OISST $\texttt{sst}$ and $\texttt{anom}$ variables~\cite{reynolds2007oisst}
and preserve a consistent grid ordering after regional cropping. The anomaly
sequence is supplied as an independent observation field. Tropical-cyclone
records from IBTrACS~\cite{knapp2010ibtracs} are filtered within the target
region and a $3^\circ$ buffer using a 34-kt wind threshold. For each forecast
date, we retain the observation with the maximum sustained wind and its location
and intensity. ENSO context is derived from monthly Ni\~no3.4 anomalies relative
to a climatological reference period. Regional circulation comes from the ERA5
$850$-hPa wind field~\cite{hersbach2020era5}. We first compute cosine-latitude-
weighted zonal and meridional means, then derive the resultant wind speed and
meteorological direction:
\begin{equation}
 \begin{aligned}
 \bar u&=\frac{\sum_j u_j\cos\varphi_j}{\sum_j\cos\varphi_j},\qquad
 \bar v=\frac{\sum_j v_j\cos\varphi_j}{\sum_j\cos\varphi_j},\\
 s&=\sqrt{\bar u^2+\bar v^2}.
 \end{aligned}
 \label{eq:wind}
\end{equation}

\paragraph{Unified forecasting prompt.}
Following the organization of text-conditioned time-series generation~\cite{wang2024news},
we construct
$q_{t,i}=\operatorname{Template}\!\left(\mathbf X_t[:,i],\mathbf D_t[:,i],
 \mathbf c_i,t,L,H,E_t,K\right)$, where $E_t$ is the organized environmental context. The instruction specifies
the target cell and historical SST, the input contains the forecast task, date
range, anomaly sequence, environmental context, and static ocean knowledge,
and the response is the comma-separated future SST sequence. During training,
the ground-truth response follows the prompt; during inference, the response is
generated autoregressively. The environmental context describes the target date,
whereas the static text provides regional background on persistence, mixing,
advection, upwelling, and air--sea exchange. Both are tokenized together with
the numerical history.

\subsection{Static and dynamic spatial graphs}
\label{sec:graph_construction}

Both graphs use the same grid nodes but encode different relations. Let
$\mathcal R(\mathbf A)$ denote row normalization with zero-sum rows preserved.
Land rows and columns are zeroed before normalization.

\paragraph{Static geographic graph.}
For great-circle distance $d_{ij}$, the geographic edge weight is
\begin{equation}
 \begin{aligned}
 \widetilde A^{s}_{ij}&=m_i m_j\mathbf 1[i\ne j]\mathbf 1[d_{ij}<r_s]
 \exp(-d_{ij}/\ell_s),\\
 \mathbf A^s&=\mathcal R(\widetilde{\mathbf A}^{s}).
 \end{aligned}
 \label{eq:static_graph}
\end{equation}
We use $\ell_s=111$ km and $r_s=1.5\ell_s$. Static node features
$\mathbf F^s\in\mathbb R^{N\times4}$ contain normalized latitude, normalized
longitude, climatological SST mean, and climatological SST standard deviation.

\paragraph{Historical SST correlation graph.}
Geographic distance alone cannot capture synchronous changes in the current
history window. We therefore compute Pearson correlations among complete,
variable ocean-node histories. Self-connections and non-positive correlations
are removed, and at most $k=10$ positive neighbors are retained per source node:
\begin{equation}
 A^{x,t}_{ij}=\begin{cases}
 \displaystyle\frac{\exp(\rho_{ij}^{(t)})}{\sum_{u\in\mathcal N_i^+(t)}\exp(\rho_{iu}^{(t)})},
 &j\in\mathcal N_i^+(t),\\[4pt]
 0,&\text{otherwise}.
 \end{cases}
 \label{eq:sst_graph}
\end{equation}

\paragraph{Localized tropical-cyclone graph.}
For a storm observation $o$ with center $\mathbf c_o$ and sustained wind $v_o$,
we define an intensity score and a finite-support influence kernel:
\begin{equation}
 \begin{aligned}
 a_o&=\operatorname{clip}(v_o/v_{99},0,1),\\
 R_o&=111\operatorname{clip}\!\left(2\sqrt{v_o/64},1.5,5\right)\ \mathrm{km},\\
 \sigma_o^2&=-R_o^2/(2\log0.1),\\
 s_i^o&=m_i a_o\exp\!\left(-\frac{d_{io}^2}{2\sigma_o^2}\right)\mathbf1[d_{io}\le R_o].
 \end{aligned}
 \label{eq:typhoon_score}
\end{equation}
Here $v_{99}$ is the 99th percentile of valid storm winds, with 130 kt used
when no valid event wind is available. The event affinity is
\begin{equation}
 \begin{aligned}
 B^o_{ij}&=\mathbf1[i\ne j]\sqrt{s_i^o s_j^o}
 \exp\!\left(-\frac{d_{ij}^2}{2\sigma_o^2}\right),\\
 \mathbf A^{e,o}&=\operatorname{diag}(\mathbf s^o)\mathcal R(\mathbf B^o).
 \end{aligned}
 \label{eq:event_graph}
\end{equation}
For multiple observations, we take the element-wise maximum, and use a zero
event graph when no storm is active. The dynamic adjacency is $\mathbf A^d_t=\mathcal R\!\left(0.6\mathbf A^{x,t}+0.4\mathbf A^{e,t}\right)$.
ENSO and regional circulation are supplied as text and explanation context;
they are not additional edge types in $\mathbf A^d_t$.

\paragraph{Dynamic node features.}
The dynamic feature matrix $\mathbf F^d_t\in\mathbb R^{N\times7}$ contains
the full-window SST mean and standard deviation, linear trend, recent seven-step
mean, full-window anomaly mean, recent anomaly mean, and SST range. Features
are standardized across ocean nodes for each date, while land features remain
zero. Node features describe the current state of each cell, while the dynamic
graph determines how these states are aggregated.

\subsection{Dual-GNN spatial encoding}
\label{sec:dual_gcn}

Geographic background and date-specific regional state evolve at different rates.
We therefore use two independent two-layer GCNs. For branch $b\in\{s,d\}$ with
features $\mathbf F^b$ and adjacency $\mathbf A^b$,
\begin{align}
 \mathbf U^b&=\operatorname{Dropout}_{0.1}\!\left[\operatorname{ReLU}\!\left(\operatorname{BN}
 (\mathbf A^b\mathbf F^b\mathbf W^b_1+\mathbf b^b_1)\right)\right],\nonumber\\
 \mathbf H^b&=\mathbf A^b\mathbf U^b\mathbf W^b_2+\mathbf b^b_2
 +\mathbf F^b\mathbf W^b_r+\mathbf b^b_r.
 \label{eq:gcn}
\end{align}
Both hidden layers have width32. The static branch outputs16 dimensions and
the dynamic branch outputs32 dimensions. After message passing over the entire
region, we concatenate the two representations and select the target cell:
\begin{equation}
 \mathbf H_t=[\mathbf H^s\Vert\mathbf H^d_t]\in\mathbb R^{N\times48},\qquad
 \mathbf h_{t,i}=\mathbf H_t[i,:]\in\mathbb R^{48}.
 \label{eq:node_selection}
\end{equation}

\subsection{Hierarchical spatial-prefix fusion}
\label{sec:prefix_fusion}

The joint spatial vector and the LLM input have different dimensions. We map
the target representation to $P=32$ continuous prefix tokens, with intermediate
token width $d_m=64$ and LLM hidden width $d_{\mathrm{LM}}$:
\begin{align}
 \mathbf u_{t,i}&=\operatorname{ReLU}(\mathbf W_1\mathbf h_{t,i}+\mathbf b_1)\in\mathbb R^{128},\nonumber\\
 \mathbf V_{t,i}&=\operatorname{reshape}_{P\times d_m}\!\left[\operatorname{ReLU}(\mathbf W_2\mathbf u_{t,i}+\mathbf b_2)\right],\nonumber\\
 \mathbf Z_{t,i}&=\alpha(\mathbf V_{t,i}\mathbf W_3+\mathbf b_3)\in\mathbb R^{P\times d_{\mathrm{LM}}}.
 \label{eq:prefix}
\end{align}
The third linear layer shares parameters across prefix positions. For Qwen2.5-7B,
$d_{\mathrm{LM}}=3584$, yielding $48\rightarrow128\rightarrow32\times64\rightarrow32\times3584$.
The projector contains 503,425 trainable parameters, compared with 5,619,713
for a direct $48\rightarrow32\times3584$ mapping, a reduction of approximately
91\% for this comparison. The learnable scalar $\alpha$ is initialized as 1 and
adjusts amplitude during training.

Let $\mathbf E_{t,i}$ denote the token embeddings of the prompt. The LLM input is $\mathbf S_{t,i}=[\mathbf Z_{t,i};\mathbf E_{t,i}]$, where the semicolon denotes concatenation along the sequence dimension. Each
sample receives only the target cell's prefix, so the prefix length does not
scale with the number of regional nodes.

\subsection{Autoregressive training and forecasting}
\label{sec:training}

The ground-truth numerical sequence is appended to the prompt as the response.
We train the model with the standard autoregressive language-model objective,
using supervised fine-tuning to predict the next response token from the prompt
and preceding tokens:
\begin{equation}
 \mathcal L=-\frac{1}{\sum_{(t,i),j}a_{t,i,j}}
 \sum_{(t,i),j}a_{t,i,j}\log p_\theta(w_{t,i,j}\mid\mathbf Z_{t,i},w_{t,i,<j}).
 \label{eq:loss}
\end{equation}
The parameters of the dual GNNs, prefix projector, prefix scale, and LoRA
adapters are jointly optimized. At inference, the numerical response is generated
autoregressively under the same conditioning sequence. Qwen2.5-7B is loaded in
8-bit precision with BF16 computation, and rank-8 LoRA adapters with scale 16 are
attached to the attention and feed-forward projections~\cite{hu2022lora}.

\begin{table*}[!t]
\caption{SST forecasting results over ten prediction steps. \textbf{Bold} entries mark the best results.}
\label{tab:main_results}
\centering
\scriptsize
\setlength{\tabcolsep}{1.6pt}
\resizebox{0.78\textwidth}{!}{%
\begin{tabular}{ccccccccccccc}
\toprule
\multirow{2}{*}{\textbf{Model}} & \multirow{2}{*}{\textbf{Metric}} & \multicolumn{10}{c}{\textbf{Days}} & \multirow{2}{*}{\textbf{Mean}}\\
\cmidrule(lr){3-12}
 & & 1 & 2 & 3 & 4 & 5 & 6 & 7 & 8 & 9 & 10 & \\
\midrule
\multirow{3}{*}{\textbf{Ours}} & RMSE & \textbf{.212} & .322 & .381 & .424 & .457 & .485 & \textbf{.512} & \textbf{.535} & \textbf{.553} & \textbf{.566} & \textbf{.445}\\
 & MAE & \textbf{.131} & \textbf{.216} & \textbf{.269} & \textbf{.306} & \textbf{.335} & \textbf{.358} & \textbf{.378} & \textbf{.396} & \textbf{.413} & \textbf{.427} & \textbf{.323}\\
 & $\Rtwo$ & \textbf{.991} & .979 & .970 & .963 & \textbf{.956} & \textbf{.950} & \textbf{.943} & \textbf{.937} & \textbf{.931} & \textbf{.927} & \textbf{.955}\\
\midrule
\multirow{3}{*}{PCFNet} & RMSE & .273 & \textbf{.294} & \textbf{.352} & \textbf{.401} & \textbf{.444} & \textbf{.484} & .514 & .545 & .574 & .597 & .448\\
 & MAE & .207 & .223 & .269 & .307 & .340 & .372 & .395 & .419 & .444 & .464 & .344\\
 & $\Rtwo$ & .983 & \textbf{.980} & \textbf{.972} & \textbf{.964} & .956 & .948 & .941 & .935 & .928 & .922 & .953\\
\midrule
\multirow{3}{*}{CoTCN} & RMSE & .281 & .336 & .382 & .430 & .477 & .515 & .540 & .568 & .588 & .602 & .472\\
 & MAE & .218 & .261 & .298 & .337 & .375 & .405 & .425 & .446 & .461 & .469 & .370\\
 & $\Rtwo$ & .982 & .974 & .967 & .958 & .949 & .941 & .935 & .929 & .924 & .921 & .948\\
\midrule
\multirow{3}{*}{PANN} & RMSE & .297 & .305 & .384 & .411 & .484 & .529 & .548 & .578 & .590 & .619 & .475\\
 & MAE & .239 & .264 & .295 & .349 & .382 & .430 & .453 & .472 & .487 & .493 & .386\\
 & $\Rtwo$ & .981 & .971 & .967 & .956 & .952 & .943 & .936 & .920 & .920 & .915 & .946\\
\midrule
\multirow{3}{*}{EarthFarseer} & RMSE & .312 & .374 & .424 & .478 & .530 & .572 & .600 & .631 & .653 & .669 & .524\\
 & MAE & .243 & .290 & .331 & .374 & .416 & .450 & .472 & .496 & .512 & .521 & .411\\
 & $\Rtwo$ & .978 & .968 & .959 & .949 & .937 & .927 & .920 & .912 & .906 & .903 & .936\\
\midrule
\multirow{3}{*}{Swin Transformer} & RMSE & .404 & .434 & .487 & .511 & .543 & .565 & .586 & .604 & .633 & .639 & .541\\
 & MAE & .331 & .355 & .400 & .415 & .439 & .457 & .472 & .486 & .511 & .515 & .438\\
 & $\Rtwo$ & .963 & .957 & .946 & .941 & .934 & .929 & .924 & .920 & .912 & .911 & .934\\
\midrule
\multirow{3}{*}{PredRNN} & RMSE & .449 & .483 & .541 & .567 & .603 & .628 & .651 & .671 & .703 & .710 & .601\\
 & MAE & .368 & .394 & .444 & .461 & .488 & .507 & .524 & .540 & .568 & .572 & .487\\
 & $\Rtwo$ & .954 & .947 & .934 & .927 & .918 & .912 & .906 & .901 & .891 & .890 & .918\\
\midrule
\multirow{3}{*}{ConvLSTM} & RMSE & .464 & .481 & .512 & .545 & .574 & .592 & .606 & .620 & .631 & .637 & .566\\
 & MAE & .396 & .408 & .431 & .458 & .481 & .494 & .503 & .513 & .521 & .524 & .473\\
 & $\Rtwo$ & .951 & .947 & .941 & .933 & .926 & .922 & .919 & .915 & .913 & .911 & .928\\
\midrule
\multirow{3}{*}{CNN} & RMSE & .515 & .544 & .566 & .602 & .641 & .687 & .712 & .742 & .750 & .763 & .652\\
 & MAE & .427 & .447 & .461 & .489 & .520 & .556 & .574 & .597 & .604 & .615 & .529\\
 & $\Rtwo$ & .939 & .932 & .927 & .918 & .908 & .895 & .888 & .878 & .877 & .873 & .904\\
\bottomrule
\end{tabular}}
\end{table*}

\subsection{Post-hoc factor-direction explanation}
\label{sec:explanation}

The explanation module reads the forecast, historical SST, and environmental
records after generation. For a temperature sequence $\mathbf v$, we compute a
linear slope and classify its direction with threshold $\tau$:
\begin{equation}
 \begin{aligned}
 \beta(\mathbf v)&=\frac{\sum_j(j-\bar j)(v_j-\bar v)}{\sum_j(j-\bar j)^2},\\
 \delta(\mathbf v)&=\begin{cases}+1,&\beta(\mathbf v)>\tau,\\-1,&\beta(\mathbf v)<-\tau,\\0,&\text{otherwise}.
 \end{cases}
  \end{aligned}
 \label{eq:direction}
\end{equation}
The current rule activates four factors: tropical cyclone influence, ENSO state,
regional wind, and historical SST trend. It then matches the active factor
directions and forecast direction against 52 explicit knowledge entries. Neutral
forecasts receive a neutral explanation, while non-neutral forecasts without a
supporting active factor receive a conflict explanation. Otherwise, the module
returns the matched knowledge identifier, mechanism description, and references.

\begin{table*}[!t]
\caption{SST forecasting results across four LLM backbones. \textbf{Bold}
values mark the best result for each metric and lead; ties at the displayed
precision are also bold.}
\label{tab:backbone_comparison}
\centering
\scriptsize
\setlength{\tabcolsep}{1.6pt}
\resizebox{0.78\textwidth}{!}{%
\begin{tabular}{ccccccccccccc}
\toprule
\multirow{2}{*}{\textbf{Model}} & \multirow{2}{*}{\textbf{Metric}} & \multicolumn{10}{c}{\textbf{Days}} & \multirow{2}{*}{\textbf{Mean}}\\
\cmidrule(lr){3-12}
 & & 1 & 2 & 3 & 4 & 5 & 6 & 7 & 8 & 9 & 10 & \\
\midrule
\multirow{3}{*}{Qwen2.5-7B-Instruct} & RMSE & .212 & .322 & .381 & .424 & .457 & .485 & \textbf{.512} & \textbf{.535} & \textbf{.553} & \textbf{.566} & .445\\
 & MAE & .131 & .216 & .269 & .306 & .335 & \textbf{.358} & \textbf{.378} & \textbf{.396} & \textbf{.413} & \textbf{.427} & \textbf{.323}\\
 & $\Rtwo$ & .991 & .979 & .970 & \textbf{.963} & \textbf{.956} & \textbf{.950} & \textbf{.943} & \textbf{.937} & \textbf{.931} & \textbf{.927} & \textbf{.955}\\
\midrule
\multirow{3}{*}{Llama-3.1-8B-Instruct} & RMSE & \textbf{.170} & \textbf{.271} & \textbf{.341} & \textbf{.397} & \textbf{.444} & \textbf{.483} & .517 & .546 & .572 & .595 & \textbf{.434}\\
 & MAE & \textbf{.115} & \textbf{.194} & \textbf{.251} & \textbf{.297} & \textbf{.334} & .366 & .393 & .417 & .439 & .458 & .326\\
 & $\Rtwo$ & \textbf{.993} & \textbf{.982} & \textbf{.971} & .961 & .952 & .943 & .934 & .927 & .920 & .913 & .950\\
\midrule
\multirow{3}{*}{Llama-2-7b-chat-hf} & RMSE & .171 & .272 & .342 & .399 & .446 & .487 & .521 & .551 & .577 & .600 & .436\\
 & MAE & .116 & .195 & .252 & \textbf{.297} & .335 & .367 & .395 & .419 & .440 & .459 & .327\\
 & $\Rtwo$ & \textbf{.993} & \textbf{.982} & \textbf{.971} & .961 & .951 & .942 & .933 & .926 & .918 & .912 & .949\\
\midrule
\multirow{3}{*}{Mistral-7B-Instruct-v0.3} & RMSE & .176 & .277 & .347 & .403 & .450 & .491 & .526 & .557 & .584 & .608 & .442\\
 & MAE & .120 & .200 & .256 & .301 & .339 & .372 & .401 & .425 & .448 & .468 & .333\\
 & $\Rtwo$ & .992 & .981 & .970 & .960 & .950 & .941 & .932 & .924 & .916 & .909 & .948\\
\bottomrule
\end{tabular}
}
\end{table*}

\section{Experiments}
\label{sec:experiments}


\subsection{Dataset and evaluation setting}
\label{sec:experimental_setup}

\textbf{Dataset:} We study SST forecasting in the South China Sea using $0.25^\circ$ OISST data.
The region is represented by a $64\times64$ regular grid, with scores computed
only on ocean cells. The data span 1 September 1981 to 16 February 2025 and are
partitioned chronologically into training, validation, and testing periods. Each
sample uses ten historical observations to predict the next ten steps, and no
window crosses a partition boundary.

\textbf{Evaluation setting:} Training and validation samples are drawn from valid windows in their respective
periods with coverage across spatial regions, coastal types, environmental events,
temperature states, and trajectory trends. Future trend labels are used only to
construct training coverage and are never provided as prediction inputs. The test
set covers all valid test dates and ocean cells, following the complete evaluation
protocol used for the PCFNet comparison. Qwen2.5-7B-Instruct is loaded in 8-bit
precision with BF16 computation. LoRA uses rank 8 and scale 16; the learning rate is
$10^{-4}$ with 100 warm-up steps, one training epoch, and random seed as 42.

\label{sec:metrics}

We report RMSE, MAE, and the coefficient of determination $\Rtwo$ at each lead,
and use the arithmetic mean across the ten leads for the summary value:
\begin{equation}
\begin{aligned}
\mathrm{RMSE}_h&=\sqrt{\frac{1}{|\mathcal S_h|}
\sum_{n\in\mathcal S_h}(\widehat y_{n,h}-y_{n,h})^2},\\
\mathrm{MAE}_h&=\frac{1}{|\mathcal S_h|}
\sum_{n\in\mathcal S_h}|\widehat y_{n,h}-y_{n,h}|,\\
R_h^2&=1-\frac{\sum_{n\in\mathcal S_h}(\widehat y_{n,h}-y_{n,h})^2}
{\sum_{n\in\mathcal S_h}(y_{n,h}-\bar y_h)^2}
\end{aligned}\nonumber
\label{eq:metrics}
\end{equation}

\subsection{Main results}
\label{sec:main_results}

Table~\ref{tab:main_results} reports lead-wise results against PCFNet and the
other specialized models listed in the corresponding SST benchmark table. Compared with the best baseline PCFNet, the complete configuration reduces mean RMSE from 0.448
to 0.445 and mean MAE from 0.344 to 0.323, while increasing mean $\Rtwo$ from
0.953 to 0.955. The RMSE comparison is lead-dependent: our model performs
better at the first prediction step and at steps 7--10, whereas PCFNet attains
lower RMSE at steps 2--6. Overall, the results indicate improved mean RMSE, MAE and $\Rtwo$.


\begin{figure*}[!t]
\centering
\includegraphics[width=0.75\textwidth]{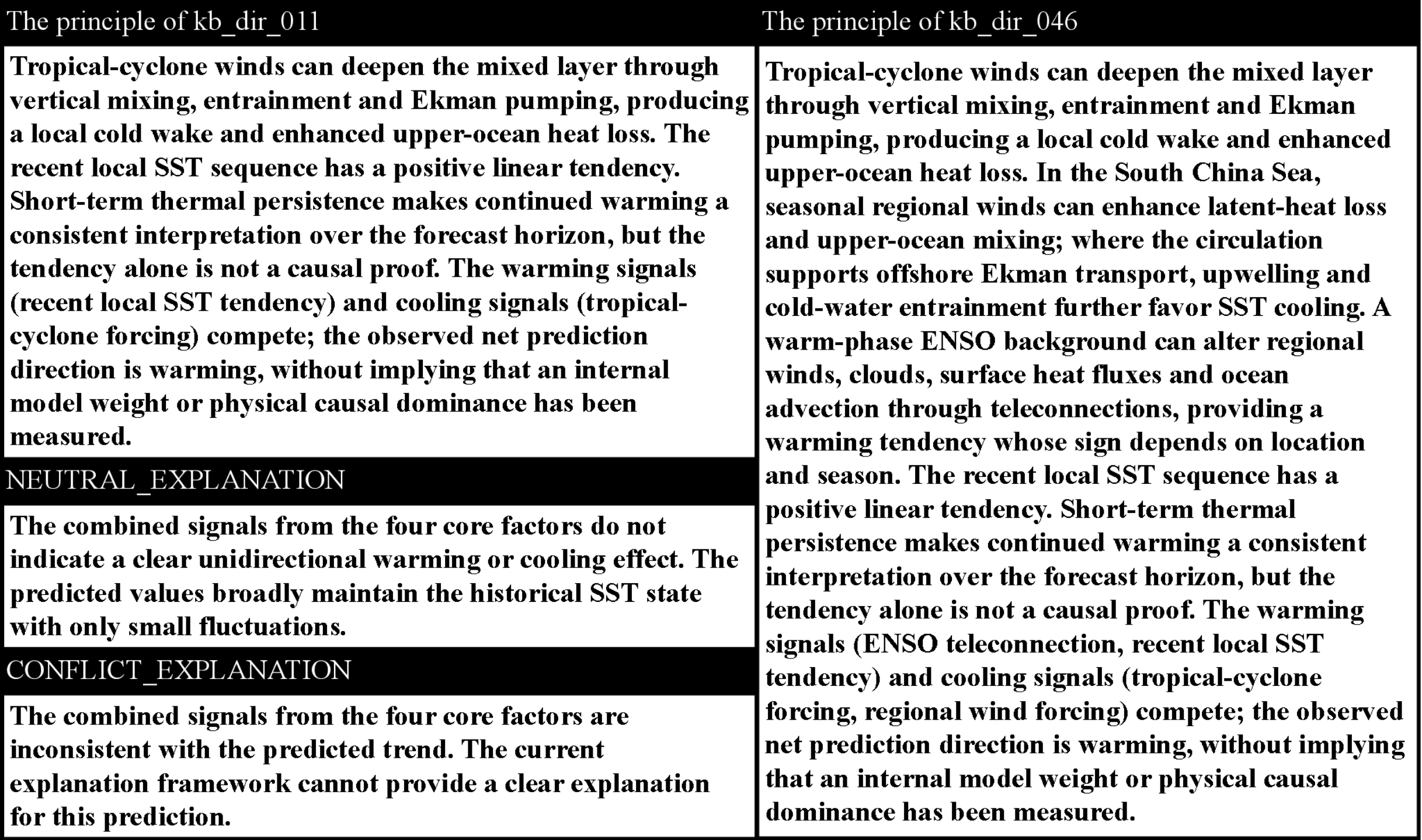}
\caption{Illustrative explanation outputs. The left column shows the principle
for \texttt{kb\_dir\_011} and the neutral and conflict fallback messages; the
right column shows the principle for \texttt{kb\_dir\_046}.}
\label{fig:explanation_examples}
\end{figure*}

\subsection{Comparison across LLM backbones}
\label{sec:backbone_comparison}

We compare the Qwen2.5-7B-Instruct backbone used in the full model with
Llama-2-7b-chat-hf, Llama-3.1-8B-Instruct, and Mistral-7B-Instruct-v0.3.
Table~\ref{tab:backbone_comparison} reports results at each forecast lead and
the arithmetic mean over ten steps. Llama-3.1-8B-Instruct achieves the lowest mean
RMSE (0.434), 0.011 below Qwen (0.445), whereas Qwen obtains the lowest mean
MAE (0.323) and the highest mean $\Rtwo$ (0.955). The RMSE comparison varies
with forecast lead: Llama-2 and Mistral have lower RMSE than Qwen at steps 1--5,
and Llama 3.1 does so at steps 1--6; Qwen has lower RMSE than all three
alternatives at steps 7--10. Thus, the backbone comparison shows a
metric- and horizon-dependent trade-off rather than a uniform improvement.

\subsection{Ablation analysis}
\label{sec:ablation}

We evaluate four variants. \texttt{no\_static} removes static ocean-knowledge
text, and \texttt{no\_dynamic} removes dynamic environmental text; both retain
the two GNN branches. \texttt{no\_knowledge} removes both text sources but
retains the spatial branch. \texttt{no\_knowledge\_no\_gnn} removes the GNN
branch entirely and therefore also has no spatial prefix input.

\begin{table}[!t]
\caption{Ablation results on the variants of our proposed framework. }
\label{tab:variants}
\centering
\footnotesize
\setlength{\tabcolsep}{2pt}
\resizebox{\linewidth}{!}{%
\begin{tabular}{ccccccc}
\toprule
Variants & \shortstack{Static-\\knowledge} & \shortstack{Dynamic-context\\text} & dual-GNN & RMSE & MAE & $\Rtwo$\\
\midrule
Full & yes & yes & yes & .445 & \textbf{.323} & \textbf{.955}\\
No static text & no & yes & yes & \textbf{.437} & .330 & .949\\
No dynamic text & yes & no & yes & .449 & .340 & .946\\
No knowledge text & no & no & yes & .456 & .347 & .944\\
No GNN & no & no & no & .464 & .354 & .943\\
\bottomrule
\end{tabular}}
\end{table}

The ablations show that the two text sources contribute complementary information.
Removing dynamic context gives RMSE 0.449, MAE 0.340, and $\Rtwo=0.946$,
whereas removing static knowledge gives RMSE 0.437, MAE 0.330, and $\Rtwo=0.949$. The complete
model has the best MAE and $\Rtwo$ among these text configurations, indicating
that combining static background with date-specific context improves overall
consistency. Removing both text sources further reduces performance to RMSE 0.456,
MAE 0.347, and $\Rtwo=0.944$.

Under the no-knowledge condition, removing the GNN branch increases RMSE from
0.456 to 0.464 and MAE from 0.347 to 0.354, while reducing $\Rtwo$ from 0.944
to 0.943. This result supports the contribution of regional spatial information
encoded by the dual GNNs. The current ablation evaluates the spatial branch as a
whole and does not separately vary the two GCNs or the prefix projector.

\subsection{Qualitative explanation examples}
\label{sec:explanation_examples}

The explanation module establishes a deterministic mapping from activated factor
directions to knowledge entries. In \texttt{kb\_dir\_011}, tropical-cyclone
influence is cooling while the recent SST trend is warming, producing a competing
two-factor pattern with a warming forecast direction. In \texttt{kb\_dir\_046},
all four factors are active: tropical cyclone and regional wind are cooling,
whereas ENSO and the recent SST trend are warming. The retrieved identifier and
references make the rule-level explanation traceable, but do not constitute an
internal attribution or a causal estimate. The module returns a neutral message for forecasts within the neutral threshold,
and a conflict message when no active factor supports a directional forecast.
Figure~\ref{fig:explanation_examples} collects the two knowledge principles and
both fallback messages discussed above.

\section{Conclusion}
\label{sec:conclusion}

We presented an LLM-based framework for regional SST forecasting that combines
textual environmental context with static and date-specific dynamic graph
representations of regional spatial state. A factorized spatial-prefix projector
maps the target-node representation to continuous tokens for autoregressive
generation, while a separate rule-based module provides source-linked post-hoc
contextualization. On South China Sea SST forecasting, the complete
configuration achieves the lowest mean RMSE, MAE and highest mean $\Rtwo$ among the
compared methods. Ablations show metric-dependent
effects of individual text components. Future work will evaluate additional
backbones and generalization across regions.

\subsection*{AI use statement}

In preparing this submission, the authors used a generative AI tool to assist with \LaTeX{} formatting, translation, and preparation of tables from existing experiment logs. 

\bibliography{references}
\bibliographystyle{IEEEtran}

\end{document}